\documentclass[10pt,twocolumn,letterpaper]{article}

\usepackage[pagenumbers]{wacv} 
\usepackage{multirow} \usepackage{placeins}
\usepackage{amsthm}
\newtheorem{theorem}{Theorem}

\definecolor{wacvblue}{rgb}{0.21,0.49,0.74}
\usepackage[pagebackref,breaklinks,colorlinks,allcolors=wacvblue]{hyperref}

\def\wacvPaperID{1959} 
\def\confName{WACV}
\def\confYear{2027}

\title{Retrieval-Augmented Initialization with Dual-Guided Inversion for Flow-Based Inverse Problem Solvers}

\author{
Hansam Cho \qquad 
Seoung Bum Kim\textsuperscript{*}\\
Department of Industrial and Management Engineering, Korea University\\
{\small\texttt{\{chosam95,sbkim1\}@korea.ac.kr}}
}

\begin{document}

\twocolumn[{
\renewcommand\twocolumn[1][]{#1}
\maketitle
\vspace{-12mm}
\begin{center}
    \centering
    \captionsetup{type=figure}
    \includegraphics[width=\linewidth]{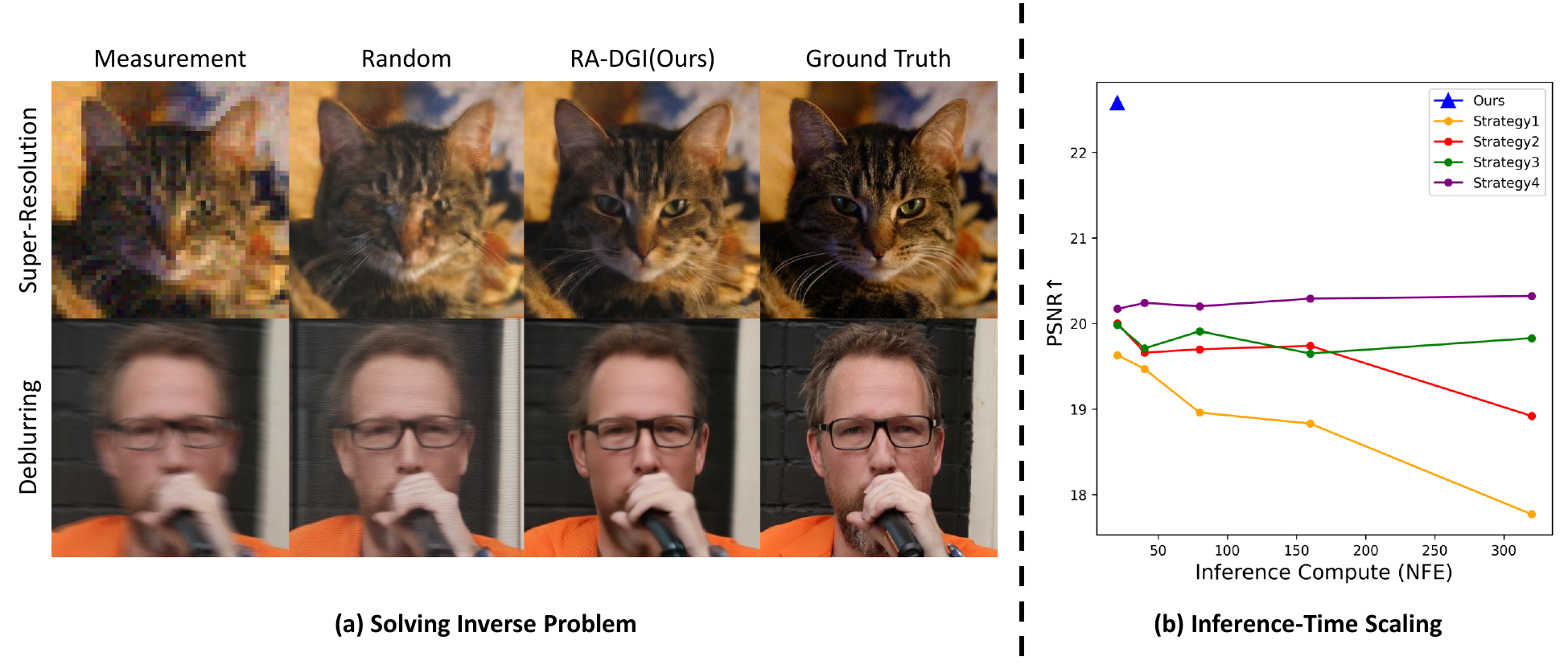}
    \vspace{-8mm}
    \caption{(a) RA-DGI produces reconstructions that are closer to the ground truth than those obtained with a random Gaussian initialization. (b) Reconstruction performance (PSNR) versus inference-time computational cost (NFE). Even with increased compute, random initialization remains inferior to RA-DGI.}
    \label{fig:teaser}
\end{center}
}]

\begingroup
\renewcommand{\thefootnote}{\fnsymbol{footnote}}
\footnotetext[1]{Corresponding author}
\endgroup

\vspace{-10mm}
\begin{abstract}
Flow models have recently emerged as powerful generative priors for solving inverse problems. Despite their success, most existing flow-based inverse problem solvers initialize the reconstruction process with random Gaussian noise. However, recent studies in image generation have shown that the choice of initial noise can significantly influence generation outcomes. The role of initialization as a distinct design choice in flow-based inverse problem solving remains comparatively underexplored. In this work, we systematically evaluate noise initialization strategies while holding the downstream flow-based inverse solver fixed. We propose Retrieval-Augmented Initialization with Dual-Guided Inversion (RA-DGI), a method that constructs improved initial noise by leveraging retrieved proxy clean images and dual-guided inversion. Extensive experiments across multiple inverse problems demonstrate that the quality of the initial noise substantially affects reconstruction fidelity and perceptual performance. RA-DGI outperforms standard random initialization and achieves superior results compared to inference-time scaling strategies, while requiring less computational overhead. These findings highlight initialization as a critical yet previously overlooked factor in flow-based inverse problem solvers.
Code is available at the following \href{https://github.com/hansam95/RA-DGI}{link}.
\end{abstract}

\vspace{-5mm}
\section{Introduction}
\label{sec:introduction}
\vspace{-5pt}
Diffusion models have become a standard approach for image synthesis~\cite{rombach2022high, ramesh2022hierarchical, podell2024sdxl}. Beyond unconditional image generation, diffusion models have also been successfully applied to solving inverse problems, where they serve as powerful image generative priors~\cite{chung2023diffusion, kawar2022denoising, wang2023zeroshot, zhang2025improving}. Formally, given a measurement $y$ generated from an unknown clean signal $x$ through a forward degradation process, the goal is to reconstruct the clean signal $x$ that is consistent with the measurement $y$. Recently, flow-based generative models have emerged as an efficient alternative to diffusion models for image generation~\cite{esser2024scaling, wu2025qwen, lipman2023flow, liu2023flow}. As a result, there has been growing interest in developing inverse problem solvers that leverage flow models~\cite{kim2025flowdps, erbach2025solving, pourya2026flower}.

Despite these advances, most inverse problem solvers still initialize the reconstruction process with randomly sampled Gaussian noise. This design choice contrasts with recent findings in image generation, where carefully chosen initial noise has been shown to significantly improve generation fidelity~\cite{zhou2025golden,guo2024initno,eyring2024reno,eyring2025noise,xu2025good}. Although initialization has received increasing attention in image generation, its role as a distinct design choice in inverse problem solving remains comparatively underexplored.

\vspace{-0.5mm}
To address this limitation, we propose a new method, Retrieval-Augmented Initialization with Dual-Guided Inversion (RA-DGI), to improve the initial noise for flow-based inverse problem solvers. First, given a degraded measurement, we retrieve a proxy clean sample from the dataset that is similar in the measurement space. This retrieved proxy clean sample serves as a more suitable starting point for inversion. Second, during inversion, we consider two complementary principles: (1) the encoded latent should remain close to the Gaussian prior distribution and (2) measurement consistency must be continuously enforced throughout the inversion process. By jointly satisfying these constraints, we obtain an initialization that is both distributionally aligned and measurement-aware.

\vspace{-0.5mm}
We demonstrate that the choice of initialization significantly affects reconstruction fidelity. As shown in Fig.~\ref{fig:teaser}(a), our method produces substantially better reconstructions than random Gaussian initialization.  Prior work has shown that inference-time scaling can significantly improve the performance of diffusion models~\cite{ma2025scaling}. Motivated by this observation, we investigate whether a similar strategy is effective for inverse problems. Fig.~\ref{fig:teaser}(b) plots reconstruction performance (PSNR, higher is better) against inference-time computational cost. Even when additional computation is allocated to random initialization, the resulting performance remains inferior to our method. These results suggest that improving initialization is more effective and computationally efficient than simply increasing the amount of inference-time compute.

\vspace{-0.5mm}
Our contributions are as follows:
\begin{itemize}
    \item {We propose RA-DGI to construct measurement-aware initial noise.}
    \item {We demonstrate that the proposed initialization outperforms random initialization and existing baselines in terms of reconstruction fidelity and perceptual quality.}
    \item {We show that our method is more computationally efficient than inference-time scaling strategies, achieving superior performance with lower computational cost.}
\end{itemize}
\section{Related Work}
\label{sec:related work}

\subsection{Solving Inverse Problems with Diffusion/Flow Models}
Inverse problems aim to recover a clean signal $x$ from degraded measurements $y$, typically modeled as $y = Ax + n$, where $A$ is a known measurement operator and $n$ denotes measurement noise. Recent approaches address this ill-posed problem using diffusion priors learned from large-scale image datasets~\cite{chung2023diffusion,kawar2022denoising,wang2023zeroshot,zhang2025improving}.

Diffusion-based inverse solvers guide the reverse process using the measurement-conditioned score $\nabla_{x_t}\log p(x_t \mid y)$, which can be decomposed by Bayes' theorem as
\[
\nabla_{x_t}\log p(x_t \mid y)
=
\nabla_{x_t}\log p(x_t)
+
\nabla_{x_t}\log p(y \mid x_t).
\]
The first term represents the learned generative prior, while the second enforces measurement consistency. Existing solvers mainly differ in how they approximate and incorporate this measurement-likelihood term~\cite{daras2024survey}.

Existing diffusion-based inverse solvers enforce measurement consistency either through likelihood gradients derived from the forward model~\cite{chung2023diffusion,kawar2022denoising,wang2023zeroshot,zhang2025improving} or through variational approximations of the posterior $p(x \mid y)$~\cite{mardani2024a,feng2023score,alkan2023variational}. Flow-based approaches extend this principle by incorporating measurement guidance into probability-flow dynamics~\cite{kim2025flowdps,pourya2026flower,erbach2025solving}. However, these flow-based solvers typically initialize reconstruction with random Gaussian noise. RA-DGI instead focuses on constructing a more informative initial latent and is complementary to existing flow-based solvers.

\subsection{Noise Initialization and Source Optimization}
Recent studies have shown that initial noise substantially affects the quality and prompt alignment of diffusion-based text-to-image generation~\cite{zhou2025golden,guo2024initno,eyring2024reno,eyring2025noise,xu2025good}. INITNO~\cite{guo2024initno} optimizes initial noise based on attention responses during early diffusion steps, while ReNO~\cite{eyring2024reno} uses reward-based optimization to improve perceptual quality. Other approaches amortize this optimization through learned networks: NPNet~\cite{zhou2025golden} transforms random noise into prompt-aligned noise, and HyperNoise~\cite{eyring2025noise} generates improved noise distributions using a prompt-conditioned hypernetwork. However, these methods primarily target prompt-aligned image generation rather than measurement-conditioned reconstruction.

Initialization has also been studied through source-space optimization for inverse problems with pretrained flow models. D-Flow~\cite{benhamu2024dflow} optimizes the initial source point for each measurement by differentiating through the full generation trajectory. MS-Flow~\cite{denker2026trajectory} instead optimizes intermediate latent states and stitches local trajectory segments, reducing the memory and numerical-stability issues of full-trajectory differentiation. Nevertheless, both methods require iterative, instance-specific optimization. RA-DGI instead constructs a measurement-aware initial latent through retrieval and dual-guided inversion, avoiding iterative optimization of source variables for each measurement.

\section{Method}
\label{sec:method}

\subsection{Background}

\paragraph{Flow Models}
Flow models~\cite{lipman2023flow, liu2023flow} aim to transport a simple source distribution $p_1$ (typically an isotropic Gaussian) to a target data distribution $p_0$. This transport is defined through a time-dependent velocity field $v_t(x)$ that governs the following ordinary differential equation (ODE):
\vspace{-2mm}
\begin{equation}
\label{eq:ode}
\frac{d x_t}{d t} = v_t(x_t), 
\quad t \in [0,1].
\end{equation}
By integrating this ODE from $t=1$ to $t=0$, a sample $x_1 \sim p_1$ is deterministically transported to a data sample $x_0 \sim p_0$. Therefore, generation amounts to estimating the marginal velocity field $v_t(x_t)$. To define the velocity field, flow models first construct a probability path $p_t$ that connects $p_1$ and $p_0$. A common choice is the linear interpolation path:
\vspace{-3mm}
\begin{equation}
x_t = a_t \boldsymbol{x}_0 + b_t x_1, \quad x_t \sim p_t,
\end{equation}
where $x_0 \sim p_0$ and $x_1 \sim p_1$, and typically $a_t = 1 - t$ and $b_t = t$.  However, the true marginal velocity field corresponding to this path is generally intractable. To address this issue, flow models are trained using the conditional flow matching (CFM) objective~\cite{lipman2023flow}:
\vspace{-2mm}
\begin{equation}
\mathcal{L}_{\mathrm{CFM}} 
= 
\mathbb{E}_{t, x_0, x_1}
\left[
\left\|
v_\theta(x_t) - v_t(x_t \mid x_0)
\right\|^2
\right],
\end{equation}
where $v_t(x_t \mid x_0)$ denotes the conditional velocity field associated with the chosen probability path. Although the training target is the conditional velocity field, minimizing this objective yields a model $v_\theta(x_t, t)$ that approximates the true marginal velocity field~\cite{lipman2023flow}. As a result, the learned vector field defines a probability flow that transports $p_1$ to $p_0$ during inference. In practice, new samples are generated by numerically solving the ODE~\ref{eq:ode} from $t=1$ to $t=0$. Using a first-order Euler discretization, the update rule is given by
\begin{equation}
x_{t-\Delta t} = x_t - \Delta t \, v_t(x_t),
\end{equation}
which progressively transforms an initial Gaussian sample into a data sample through step-wise integration.
\vspace{-3mm}
\paragraph{Inversion}
Since the ODE defines a deterministic mapping between the data and latent distributions, it can be solved in both time directions. In particular, given a data sample $x_0$, we can obtain its corresponding latent representation by integrating the ODE forward from $t=0$ to $t=1$. Using a first-order Euler discretization, the update rule is given by
\vspace{-2mm}
\begin{equation}
\label{eq:euler_inv}
x_{t+\Delta t} = x_t + \Delta t \, v_t(x_t),
\end{equation}
which progressively transports the data sample toward its inverted latent code $x_1^{\mathrm{inv}}$ in the Gaussian space. Conversely, starting from $x_1^{\mathrm{inv}}$ and integrating the ODE backward from $t=1$ to $t=0$ recovers the original data sample under sufficiently accurate numerical integration~\cite{rout2025semantic}.

\begin{figure*}[ht]
    \centering
    \includegraphics[width=\linewidth]{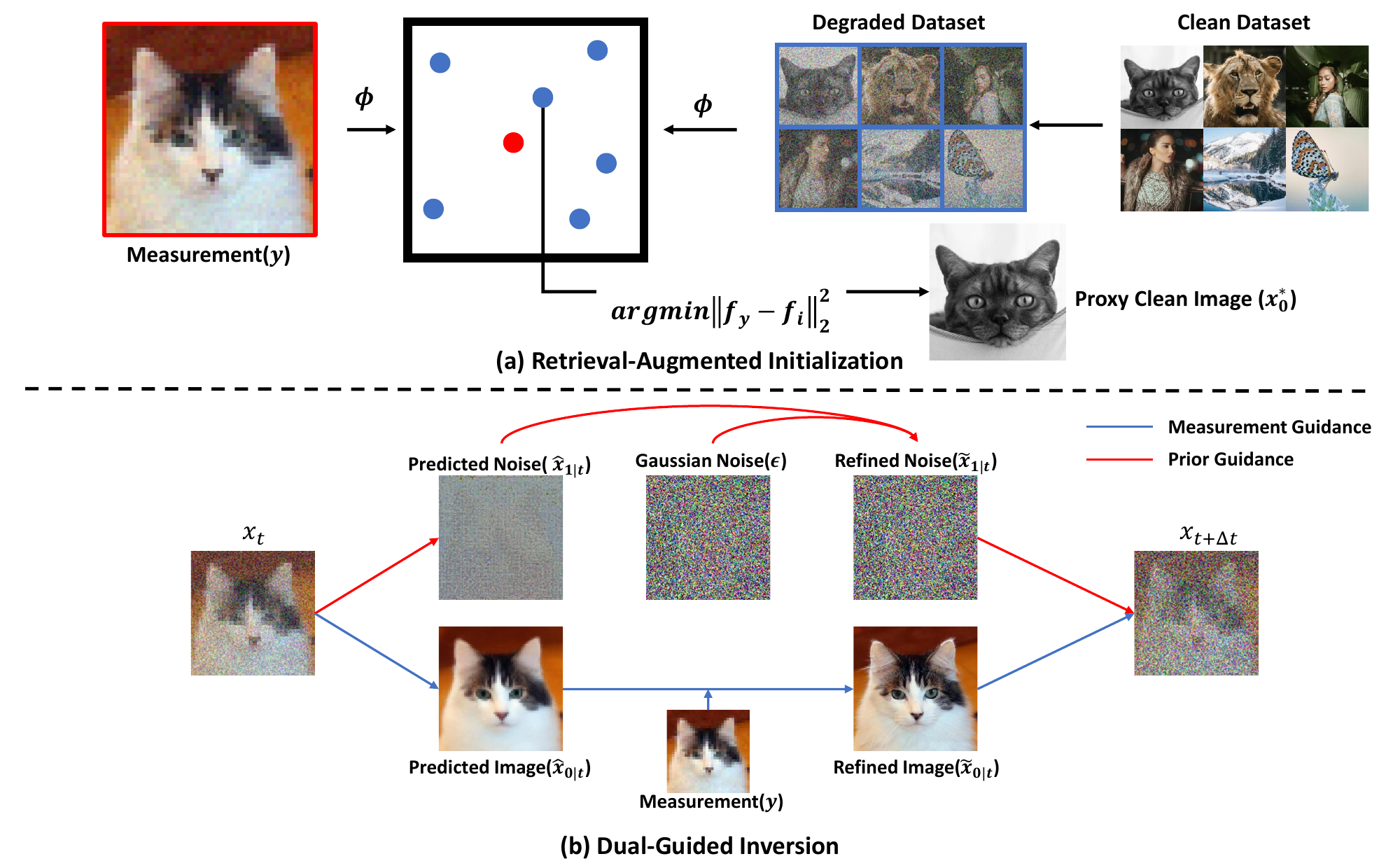}
    \caption{\textbf{Overview of RA-DGI.}
    (a) Retrieval-augmented initialization. A proxy clean image $x_0^\ast$ is retrieved from a dataset using semantic features extracted by $\phi(\cdot)$ and used as the starting point for inversion. 
    (b) Dual-guided inversion. Measurement guidance (blue) enforces measurement consistency, while prior guidance (red) applies Gaussian regularization to encourage compatibility with the source prior, jointly refining the latent trajectory.}
    \vspace{-3mm}
    \label{fig:method}
\end{figure*}
\vspace{-3mm}
\paragraph{Tweedie's Formula for Flow Models}
Tweedie's formula provides a way to estimate clean variables from intermediate states through conditional expectations. 
FlowDPS~\cite{kim2025flowdps} extends this idea to flow models and shows that both the clean data sample $x_0$ and the latent noise sample $x_1$ can be estimated from an intermediate state $x_t$ using the velocity field. Specifically, the conditional expectations of $x_0$ and $x_1$ given $x_t$ are expressed as
\vspace{-2mm}
\begin{equation}
\label{eq:Tweedie}
\begin{aligned}
\mathbb{E}[x_0 \mid x_t]
&= x_t - t\, v_t(x_t) \\
\mathbb{E}[x_1 \mid x_t]
&= x_t + (1-t)\, v_t(x_t)
\end{aligned}
\end{equation}
where $v_t(x_t)$ denotes the marginal velocity field at time $t$. 
Using these expressions, the velocity field can be decomposed as
\vspace{-2mm}
\begin{equation}
\label{eq:vel_decomp}
v_t(x_t)
=
-\mathbb{E}[x_0 \mid x_t]
+
\mathbb{E}[x_1 \mid x_t].
\end{equation}

\subsection{Retrieval-Augmented Initialization}
A straightforward strategy for obtaining an initial latent code is to directly invert the given measurement $y$.
However, flow models are trained on clean images rather than degraded measurements. As a result, directly inverting measurement often produces a latent representation that deviates from the Gaussian prior, which can lead to poor reconstruction quality~\cite{rout2025semantic}. To mitigate this issue, we instead retrieve a proxy clean sample from the dataset and use it as the starting point for inversion. Intuitively, a clean image that is consistent with the measurement can provide a more suitable initialization for inversion.

Following the standard formulation of inverse problems, the measurement model is written as
\vspace{-2mm}
\begin{equation}
y = A x_0 + n,
\end{equation}
where $A \in \mathbb{R}^{m \times d}$ denotes the forward degradation matrix and $n \sim \mathcal{N}(0, \sigma_n^2 I_m)$ represents additive Gaussian noise with variance $\sigma_n^2$. Let $A^\dagger$ denote the pseudo-inverse of $A$, which maps measurements back to the image space. For each clean sample $x_0^i \in \mathcal{D}$, we simulate the same degradation process and map the result back to the image space:
\begin{equation}
\tilde{x}_0^i = A^\dagger A x_0^i .
\end{equation}
We then extract feature representations of each degraded-clean sample and the measurement using a pretrained encoder $\phi(\cdot)$:
\vspace{-2mm}
\begin{equation}
f_i = \phi(\tilde{x}_0^i), 
\qquad 
f_y = \phi(A^\dagger y).
\end{equation}
The proxy sample is selected by minimizing the feature distance:
\vspace{-2mm}
\begin{equation}
x_0^\ast
=
\arg\min_{x_0^i \in \mathcal{D}}
\| f_y - f_i \|_2^2 .
\end{equation}
The retrieved proxy clean image $x_0^\ast$ serves as the starting point for inversion. Inverting from $x_0^\ast$ yields a latent code that more closely aligns with the Gaussian prior, thereby improving reconstruction performance. Fig.~\ref{fig:method}(a) illustrates the overall process of retrieval-augmented initialization.

\subsection{Dual-Guided Inversion}
\vspace{-1mm}
\paragraph{Measurement Guidance}
Because the inversion trajectory is initialized from the proxy clean image, the resulting latent code tends to reconstruct the proxy image rather than satisfy the measurement constraint $y$. Our objective, however, is to obtain a latent representation that is consistent with the measurement. To this end, we perform inversion using a measurement-conditional velocity field $v_t(x_t \mid y)$ instead of the unconditional velocity field $v_t(x_t)$. Accordingly, Eq.~\ref{eq:euler_inv} is modified as follows:
\vspace{-2mm}
\begin{equation}
\label{eq:euler_inv_measure}
x_{t+\Delta t} = x_t + \Delta t \, v_\theta(x_t \mid y).
\end{equation}
The measurement-conditional velocity field can be expressed as
\begin{equation}
v_t(x_t \mid y) = v_t(x_t) + C_1(t) \nabla_{\hat{x}_{0|t}} \log p(y \mid \hat{x}_{0|t}),
\end{equation}
where $\hat{x}_{0|t} := \mathbb{E}[x_0 \mid x_t]$ and $C_1(t)= \frac{t}{(1-t)^2}$~\cite{kim2025flowdps}. Substituting this expression into Eq.~\ref{eq:euler_inv_measure} gives
\begin{equation}
x_{t+\Delta t}
=
x_t
+
\Delta t\, v_t(x_t)
+
\Delta t\, C_1(t) \nabla_{\hat{x}_{0|t}} \log p(y \mid \hat{x}_{0|t}).
\end{equation}
Applying Tweedie's formula (Eq.~\ref{eq:Tweedie}) further expands this expression as
\vspace{-2mm}
\begin{equation}
\begin{split}
x_{t+\Delta t}
&=
(\hat{x}_{0|t}+t v_t(x_t))
+
\Delta t\, v_t(x_t) \\
&\quad
+
\Delta t\, C_1(t)
\nabla_{\hat{x}_{0|t}}
\log p(y \mid \hat{x}_{0|t}),
\end{split}
\end{equation}
Furthermore, by applying the velocity decomposition in Eq.~\ref{eq:vel_decomp}, we obtain
\vspace{-2mm}
\begin{align}
x_{t+\Delta t}
&= (\hat{x}_{0|t}+t(-\hat{x}_{0|t}+\hat{x}_{1|t}))
+ \Delta t(-\hat{x}_{0|t}+\hat{x}_{1|t}) \nonumber\\
&\quad + \Delta t\, C_1(t) \nabla_{\hat{x}_{0|t}} \log p(y \mid \hat{x}_{0|t}) \\
&= C_2(t) \hat{x}_{0|t} + C_3(t) \hat{x}_{1|t}  \nonumber\\
&\quad + \Delta t\, C_1(t) \nabla_{\hat{x}_{0|t}} \log p(y \mid \hat{x}_{0|t}) \\
&= C_2(t)\tilde{x}_{0|t}
+ C_3(t) \hat{x}_{1|t}. \label{eq:inv_step}
\end{align}
where
\[
\hat{x}_{1|t} := \mathbb{E}[x_1 \mid x_t], 
C_2(t) = 1 - t - \Delta t,
C_3(t) = t + \Delta t,
\]
and
\vspace{-2mm}
\[
\tilde{x}_{0|t}
=
\hat{x}_{0|t}
+
\frac{\Delta t\, C_1(t)}{C_2(t)}
\nabla_{\hat{x}_{0|t}} \log p(y \mid \hat{x}_{0|t}).
\]
By continuously incorporating this measurement-consistency term into the inversion dynamics, the latent trajectory is guided toward regions that are consistent with the measurement.
\vspace{-4mm}
\paragraph{Prior guidance}
Focusing solely on measurement guidance may cause the encoded latent to deviate from the Gaussian prior. To address this issue, we further modify the measurement-guided inversion step. The term $\hat{x}_{1|t}$ represents the predicted noise inferred from the latent state $x_t$, which implicitly contains structural information derived from the measurement. Consequently, directly using $\hat{x}_{1|t}$ in the inversion dynamics may bias the latent toward measurement-specific structures rather than maintaining consistency with the Gaussian prior.

To mitigate this effect, we introduce a noise regularization mechanism that injects Gaussian noise to weaken the structural information contained in $\hat{x}_{1|t}$. Specifically, we replace $\hat{x}_{1|t}$ with the following noise-refined representation:
\vspace{-2mm}
\begin{equation}
\tilde{x}_{1|t} = \sqrt{1-t}\, \hat{x}_{1|t} + \sqrt{t}\, \epsilon_t,
\end{equation}
where $\epsilon_t \sim \mathcal{N}(0,I)$ denotes a standard Gaussian sample that is independently resampled at every inversion step. This stochastic mixing acts as a regularizer that reduces persistent measurement-specific structure and encourages compatibility with the Gaussian source prior. With this modification, the final inversion update in Eq.~\ref{eq:inv_step} becomes
\vspace{-2mm}
\begin{equation}
x_{t+\Delta t}
=
C_2(t)\tilde{x}_{0|t}
+
C_3(t)\tilde{x}_{1|t}.
\end{equation}
Fig.~\ref{fig:method}(b) illustrates the overall process of dual-guided inversion.
\vspace{-2mm}
\section{Experiments}
\label{sec:experiments}

\subsection{Experimental Settings}

\paragraph{Inverse Problems}
We set the number of function evaluations (NFE) to 10 for all experiments. We evaluate our method on four linear inverse problems: super-resolution, deblurring, denoising, and compressed sensing. For super-resolution, we consider a $\times16$ downscaling setting, while for deblurring, we apply motion blur with a kernel size of 81. For denoising, additive Gaussian noise with a standard deviation of $\sigma = 0.7$ is used. For compressed sensing, we adopt a block-wise compressed sensing scheme with a block size of $32\times32$ and a measurement ratio of $0.1$. For all inverse problems except denoising, measurement noise with a standard deviation of $\sigma_y = 0.05$ is added.

\begin{table*}[t]
\centering
\scriptsize
\setlength{\tabcolsep}{2pt}
\renewcommand{\arraystretch}{1.15}
\resizebox{\linewidth}{!}{
\begin{tabular}{c|cccc|cccc|cccc|cccc}

\toprule
\multicolumn{17}{c}{AFHQ} \\
\midrule
\multirow{2}{*}{Method}
& \multicolumn{4}{c|}{Super-Resolution}
& \multicolumn{4}{c|}{Deblurring}
& \multicolumn{4}{c|}{Denoising}
& \multicolumn{4}{c}{Compressed Sensing} \\
\cmidrule(lr){2-5} \cmidrule(lr){6-9} \cmidrule(lr){10-13} \cmidrule(lr){14-17}
& PSNR$\uparrow$ & SSIM$\uparrow$ & LPIPS$\downarrow$ & FID$\downarrow$ & PSNR$\uparrow$ & SSIM$\uparrow$ & LPIPS$\downarrow$ & FID$\downarrow$ & PSNR$\uparrow$ & SSIM$\uparrow$ & LPIPS$\downarrow$ & FID$\downarrow$ & PSNR$\uparrow$ & SSIM$\uparrow$ & LPIPS$\downarrow$ & FID$\downarrow$ \\
\midrule
Random & \shortstack{23.34\\{\scriptsize (0.050)}} & \shortstack{0.477\\{\scriptsize (0.004)}} & \shortstack{\textbf{0.235}\\{\scriptsize (0.004)}} & \shortstack{\textbf{21.91}\\{\scriptsize (0.763)}} & \shortstack{22.79\\{\scriptsize (0.021)}} & \shortstack{0.464\\{\scriptsize (0.003)}} & \shortstack{\textbf{0.254}\\{\scriptsize (0.004)}} & \shortstack{24.76\\{\scriptsize (1.272)}} & \shortstack{21.92\\{\scriptsize (0.015)}} & \shortstack{0.446\\{\scriptsize (0.004)}} & \shortstack{0.301\\{\scriptsize (0.004)}} & \shortstack{31.22\\{\scriptsize (0.720)}} & \shortstack{20.25\\{\scriptsize (0.006)}} & \shortstack{0.423\\{\scriptsize (0.003)}} & \shortstack{0.353\\{\scriptsize (0.004)}} & \shortstack{41.72\\{\scriptsize (1.104)}} \\
NPNet & \shortstack{23.18\\{\scriptsize (0.067)}} & \shortstack{0.474\\{\scriptsize (0.003)}} & \shortstack{0.240\\{\scriptsize (0.007)}} & \shortstack{24.41\\{\scriptsize (1.159)}} & \shortstack{22.59\\{\scriptsize (0.061)}} & \shortstack{0.453\\{\scriptsize (0.003)}} & \shortstack{\underline{0.261}\\{\scriptsize (0.006)}} & \shortstack{26.62\\{\scriptsize (1.306)}} & \shortstack{21.93\\{\scriptsize (0.060)}} & \shortstack{\underline{0.446}\\{\scriptsize (0.003)}} & \shortstack{0.293\\{\scriptsize (0.005)}} & \shortstack{33.24\\{\scriptsize (1.330)}} & \shortstack{20.35\\{\scriptsize (0.061)}} & \shortstack{0.416\\{\scriptsize (0.003)}} & \shortstack{0.338\\{\scriptsize (0.003)}} & \shortstack{42.17\\{\scriptsize (1.227)}} \\
RF-Inversion & \shortstack{\underline{23.94}\\{\scriptsize (0.064)}} & \shortstack{\underline{0.529}\\{\scriptsize (0.004)}} & \shortstack{0.287\\{\scriptsize (0.000)}} & \shortstack{22.95\\{\scriptsize (1.104)}} & \shortstack{\underline{23.22}\\{\scriptsize (0.057)}} & \shortstack{\underline{0.486}\\{\scriptsize (0.003)}} & \shortstack{0.299\\{\scriptsize (0.004)}} & \shortstack{\textbf{23.83}\\{\scriptsize (0.693)}} & \shortstack{\underline{22.05}\\{\scriptsize (0.075)}} & \shortstack{0.406\\{\scriptsize (0.005)}} & \shortstack{\textbf{0.243}\\{\scriptsize (0.003)}} & \shortstack{\textbf{28.27}\\{\scriptsize (0.473)}} & \shortstack{\underline{22.32}\\{\scriptsize (0.053)}} & \shortstack{\underline{0.441}\\{\scriptsize (0.002)}} & \shortstack{\underline{0.267}\\{\scriptsize (0.007)}} & \shortstack{\underline{29.97}\\{\scriptsize (0.868)}} \\
RA-DGI (Ours) & \shortstack{\textbf{24.03}\\{\scriptsize (0.101)}} & \shortstack{\textbf{0.537}\\{\scriptsize (0.003)}} & \shortstack{\underline{0.236}\\{\scriptsize (0.007)}} & \shortstack{\underline{21.92}\\{\scriptsize (0.768)}} & \shortstack{\textbf{23.57}\\{\scriptsize (0.115)}} & \shortstack{\textbf{0.528}\\{\scriptsize (0.003)}} & \shortstack{0.266\\{\scriptsize (0.006)}} & \shortstack{\underline{24.69}\\{\scriptsize (0.668)}} & \shortstack{\textbf{22.85}\\{\scriptsize (0.085)}} & \shortstack{\textbf{0.475}\\{\scriptsize (0.002)}} & \shortstack{\underline{0.266}\\{\scriptsize (0.004)}} & \shortstack{\underline{28.38}\\{\scriptsize (0.286)}} & \shortstack{\textbf{22.69}\\{\scriptsize (0.025)}} & \shortstack{\textbf{0.454}\\{\scriptsize (0.000)}} & \shortstack{\textbf{0.262}\\{\scriptsize (0.001)}} & \shortstack{\textbf{29.72}\\{\scriptsize (0.229)}} \\

\midrule
\multicolumn{17}{c}{FFHQ} \\
\midrule
\multirow{2}{*}{Method}
& \multicolumn{4}{c|}{Super-Resolution}
& \multicolumn{4}{c|}{Deblurring}
& \multicolumn{4}{c|}{Denoising}
& \multicolumn{4}{c}{Compressed Sensing} \\
\cmidrule(lr){2-5} \cmidrule(lr){6-9} \cmidrule(lr){10-13} \cmidrule(lr){14-17}
& PSNR$\uparrow$ & SSIM$\uparrow$ & LPIPS$\downarrow$ & FID$\downarrow$ & PSNR$\uparrow$ & SSIM$\uparrow$ & LPIPS$\downarrow$ & FID$\downarrow$ & PSNR$\uparrow$ & SSIM$\uparrow$ & LPIPS$\downarrow$ & FID$\downarrow$ & PSNR$\uparrow$ & SSIM$\uparrow$ & LPIPS$\downarrow$ & FID$\downarrow$ \\
\midrule
Random & \shortstack{23.04\\{\scriptsize (0.006)}} & \shortstack{0.461\\{\scriptsize (0.005)}} & \shortstack{\textbf{0.210}\\{\scriptsize (0.004)}} & \shortstack{62.98\\{\scriptsize (0.693)}} & \shortstack{22.61\\{\scriptsize (0.085)}} & \shortstack{0.447\\{\scriptsize (0.003)}} & \shortstack{\underline{0.237}\\{\scriptsize (0.003)}} & \shortstack{\underline{61.18}\\{\scriptsize (0.542)}} & \shortstack{\underline{21.83}\\{\scriptsize (0.064)}} & \shortstack{\underline{0.433}\\{\scriptsize (0.001)}} & \shortstack{0.248\\{\scriptsize (0.004)}} & \shortstack{68.92\\{\scriptsize (0.730)}} & \shortstack{20.45\\{\scriptsize (0.078)}} & \shortstack{0.404\\{\scriptsize (0.002)}} & \shortstack{0.298\\{\scriptsize (0.007)}} & \shortstack{81.98\\{\scriptsize (0.779)}} \\
NPNet & \shortstack{22.28\\{\scriptsize (0.096)}} & \shortstack{0.419\\{\scriptsize (0.003)}} & \shortstack{0.246\\{\scriptsize (0.003)}} & \shortstack{79.00\\{\scriptsize (1.227)}} & \shortstack{21.96\\{\scriptsize (0.042)}} & \shortstack{0.402\\{\scriptsize (0.002)}} & \shortstack{0.271\\{\scriptsize (0.004)}} & \shortstack{79.19\\{\scriptsize (0.267)}} & \shortstack{21.66\\{\scriptsize (0.069)}} & \shortstack{0.393\\{\scriptsize (0.001)}} & \shortstack{0.247\\{\scriptsize (0.003)}} & \shortstack{78.74\\{\scriptsize (0.934)}} & \shortstack{20.67\\{\scriptsize (0.066)}} & \shortstack{0.370\\{\scriptsize (0.003)}} & \shortstack{0.283\\{\scriptsize (0.004)}} & \shortstack{90.22\\{\scriptsize (1.320)}} \\
RF-Inversion & \shortstack{\underline{23.82}\\{\scriptsize (0.057)}} & \shortstack{\underline{0.525}\\{\scriptsize (0.005)}} & \shortstack{0.246\\{\scriptsize (0.005)}} & \shortstack{\underline{58.98}\\{\scriptsize (0.473)}} & \shortstack{\underline{22.95}\\{\scriptsize (0.061)}} & \shortstack{\underline{0.470}\\{\scriptsize (0.001)}} & \shortstack{0.276\\{\scriptsize (0.004)}} & \shortstack{64.48\\{\scriptsize (0.730)}} & \shortstack{21.43\\{\scriptsize (0.115)}} & \shortstack{0.391\\{\scriptsize (0.004)}} & \shortstack{\textbf{0.197}\\{\scriptsize (0.003)}} & \shortstack{\textbf{62.17}\\{\scriptsize (0.705)}} & \shortstack{\underline{21.64}\\{\scriptsize (0.031)}} & \shortstack{\underline{0.417}\\{\scriptsize (0.003)}} & \shortstack{\underline{0.230}\\{\scriptsize (0.004)}} & \shortstack{\textbf{67.61}\\{\scriptsize (1.330)}} \\
RA-DGI (Ours) & \shortstack{\textbf{24.41}\\{\scriptsize (0.064)}} & \shortstack{\textbf{0.570}\\{\scriptsize (0.003)}} & \shortstack{\underline{0.210}\\{\scriptsize (0.003)}} & \shortstack{\textbf{53.46}\\{\scriptsize (0.509)}} & \shortstack{\textbf{23.88}\\{\scriptsize (0.025)}} & \shortstack{\textbf{0.547}\\{\scriptsize (0.000)}} & \shortstack{\textbf{0.237}\\{\scriptsize (0.001)}} & \shortstack{\textbf{57.47}\\{\scriptsize (0.462)}} & \shortstack{\textbf{22.90}\\{\scriptsize (0.053)}} & \shortstack{\textbf{0.472}\\{\scriptsize (0.003)}} & \shortstack{\underline{0.218}\\{\scriptsize (0.004)}} & \shortstack{\underline{64.39}\\{\scriptsize (0.743)}} & \shortstack{\textbf{22.43}\\{\scriptsize (0.060)}} & \shortstack{\textbf{0.433}\\{\scriptsize (0.003)}} & \shortstack{\textbf{0.224}\\{\scriptsize (0.004)}} & \shortstack{\underline{81.53}\\{\scriptsize (0.798)}} \\

\midrule
\multicolumn{17}{c}{DIV2K} \\
\midrule
\multirow{2}{*}{Method}
& \multicolumn{4}{c|}{Super-Resolution}
& \multicolumn{4}{c|}{Deblurring}
& \multicolumn{4}{c|}{Denoising}
& \multicolumn{4}{c}{Compressed Sensing} \\
\cmidrule(lr){2-5} \cmidrule(lr){6-9} \cmidrule(lr){10-13} \cmidrule(lr){14-17}
& PSNR$\uparrow$ & SSIM$\uparrow$ & LPIPS$\downarrow$ & FID$\downarrow$ & PSNR$\uparrow$ & SSIM$\uparrow$ & LPIPS$\downarrow$ & FID$\downarrow$ & PSNR$\uparrow$ & SSIM$\uparrow$ & LPIPS$\downarrow$ & FID$\downarrow$ & PSNR$\uparrow$ & SSIM$\uparrow$ & LPIPS$\downarrow$ & FID$\downarrow$ \\
\midrule
Random & \shortstack{18.88\\{\scriptsize (0.060)}} & \shortstack{0.334\\{\scriptsize (0.005)}} & \shortstack{\textbf{0.370}\\{\scriptsize (0.003)}} & \shortstack{96.61\\{\scriptsize (0.868)}} & \shortstack{18.78\\{\scriptsize (0.062)}} & \shortstack{0.344\\{\scriptsize (0.001)}} & \shortstack{\underline{0.400}\\{\scriptsize (0.004)}} & \shortstack{\underline{104.52}\\{\scriptsize (0.779)}} & \shortstack{19.10\\{\scriptsize (0.046)}} & \shortstack{\underline{0.382}\\{\scriptsize (0.005)}} & \shortstack{0.379\\{\scriptsize (0.000)}} & \shortstack{96.57\\{\scriptsize (1.330)}} & \shortstack{17.48\\{\scriptsize (0.044)}} & \shortstack{0.325\\{\scriptsize (0.003)}} & \shortstack{0.464\\{\scriptsize (0.004)}} & \shortstack{115.29\\{\scriptsize (0.163)}} \\
NPNet & \shortstack{18.81\\{\scriptsize (0.061)}} & \shortstack{0.339\\{\scriptsize (0.003)}} & \shortstack{0.410\\{\scriptsize (0.003)}} & \shortstack{109.16\\{\scriptsize (0.267)}} & \shortstack{18.90\\{\scriptsize (0.064)}} & \shortstack{0.342\\{\scriptsize (0.002)}} & \shortstack{0.414\\{\scriptsize (0.004)}} & \shortstack{109.26\\{\scriptsize (1.316)}} & \shortstack{19.14\\{\scriptsize (0.092)}} & \shortstack{0.374\\{\scriptsize (0.003)}} & \shortstack{0.391\\{\scriptsize (0.004)}} & \shortstack{99.44\\{\scriptsize (0.516)}} & \shortstack{17.33\\{\scriptsize (0.070)}} & \shortstack{0.324\\{\scriptsize (0.003)}} & \shortstack{0.467\\{\scriptsize (0.001)}} & \shortstack{115.14\\{\scriptsize (0.867)}} \\
RF-Inversion & \shortstack{\textbf{19.59}\\{\scriptsize (0.075)}} & \shortstack{\underline{0.381}\\{\scriptsize (0.004)}} & \shortstack{0.414\\{\scriptsize (0.004)}} & \shortstack{\underline{87.99}\\{\scriptsize (0.823)}} & \shortstack{\underline{19.42}\\{\scriptsize (0.115)}} & \shortstack{\underline{0.357}\\{\scriptsize (0.002)}} & \shortstack{0.453\\{\scriptsize (0.003)}} & \shortstack{112.42\\{\scriptsize (1.050)}} & \shortstack{\underline{19.60}\\{\scriptsize (0.015)}} & \shortstack{0.349\\{\scriptsize (0.002)}} & \shortstack{\textbf{0.320}\\{\scriptsize (0.003)}} & \shortstack{\textbf{88.03}\\{\scriptsize (0.404)}} & \shortstack{\underline{18.67}\\{\scriptsize (0.105)}} & \shortstack{\underline{0.326}\\{\scriptsize (0.003)}} & \shortstack{\textbf{0.394}\\{\scriptsize (0.004)}} & \shortstack{\underline{109.16}\\{\scriptsize (1.267)}} \\
RA-DGI (Ours) & \shortstack{\underline{19.42}\\{\scriptsize (0.105)}} & \shortstack{\textbf{0.392}\\{\scriptsize (0.002)}} & \shortstack{\underline{0.376}\\{\scriptsize (0.002)}} & \shortstack{\textbf{86.93}\\{\scriptsize (0.229)}} & \shortstack{\textbf{19.56}\\{\scriptsize (0.080)}} & \shortstack{\textbf{0.398}\\{\scriptsize (0.003)}} & \shortstack{\textbf{0.394}\\{\scriptsize (0.007)}} & \shortstack{\textbf{93.34}\\{\scriptsize (0.697)}} & \shortstack{\textbf{19.81}\\{\scriptsize (0.067)}} & \shortstack{\textbf{0.405}\\{\scriptsize (0.003)}} & \shortstack{\underline{0.346}\\{\scriptsize (0.003)}} & \shortstack{\underline{89.67}\\{\scriptsize (0.794)}} & \shortstack{\textbf{18.72}\\{\scriptsize (0.076)}} & \shortstack{\textbf{0.349}\\{\scriptsize (0.002)}} & \shortstack{\underline{0.398}\\{\scriptsize (0.003)}} & \shortstack{\textbf{107.66}\\{\scriptsize (1.159)}} \\

\bottomrule
\end{tabular}
}
\vspace{-2mm}
\caption{Quantitative comparison on four inverse problems across three datasets. Values are reported as the mean over three seeds, with the sample standard deviation shown in parentheses. Best and second-best mean values are shown in \textbf{bold} and \underline{underline}, respectively.}
\vspace{-3mm}
\label{tab:quant_result}
\end{table*}

\vspace{-5mm}
\paragraph{Datasets and Metrics}
We conduct experiments on three datasets: AFHQ~\cite{choi2020stargan}, FFHQ~\cite{karras2019style}, and DIV2K~\cite{Agustsson_2017_CVPR_Workshops}. From AFHQ, we use 1,000 images from the validation set, and from FFHQ, we select 1,000 images. From DIV2K, we use 800 high-resolution images from the validation set. For quantitative evaluation, we report PSNR and SSIM to measure pixel-level reconstruction accuracy. To assess perceptual quality, we use LPIPS~\cite{zhang2018unreasonable} as a perceptual similarity metric and FID~\cite{heusel2017gans} to evaluate distributional alignment with the ground-truth data distribution.

\begin{figure*}[ht]
    \centering
    \includegraphics[width=\linewidth]{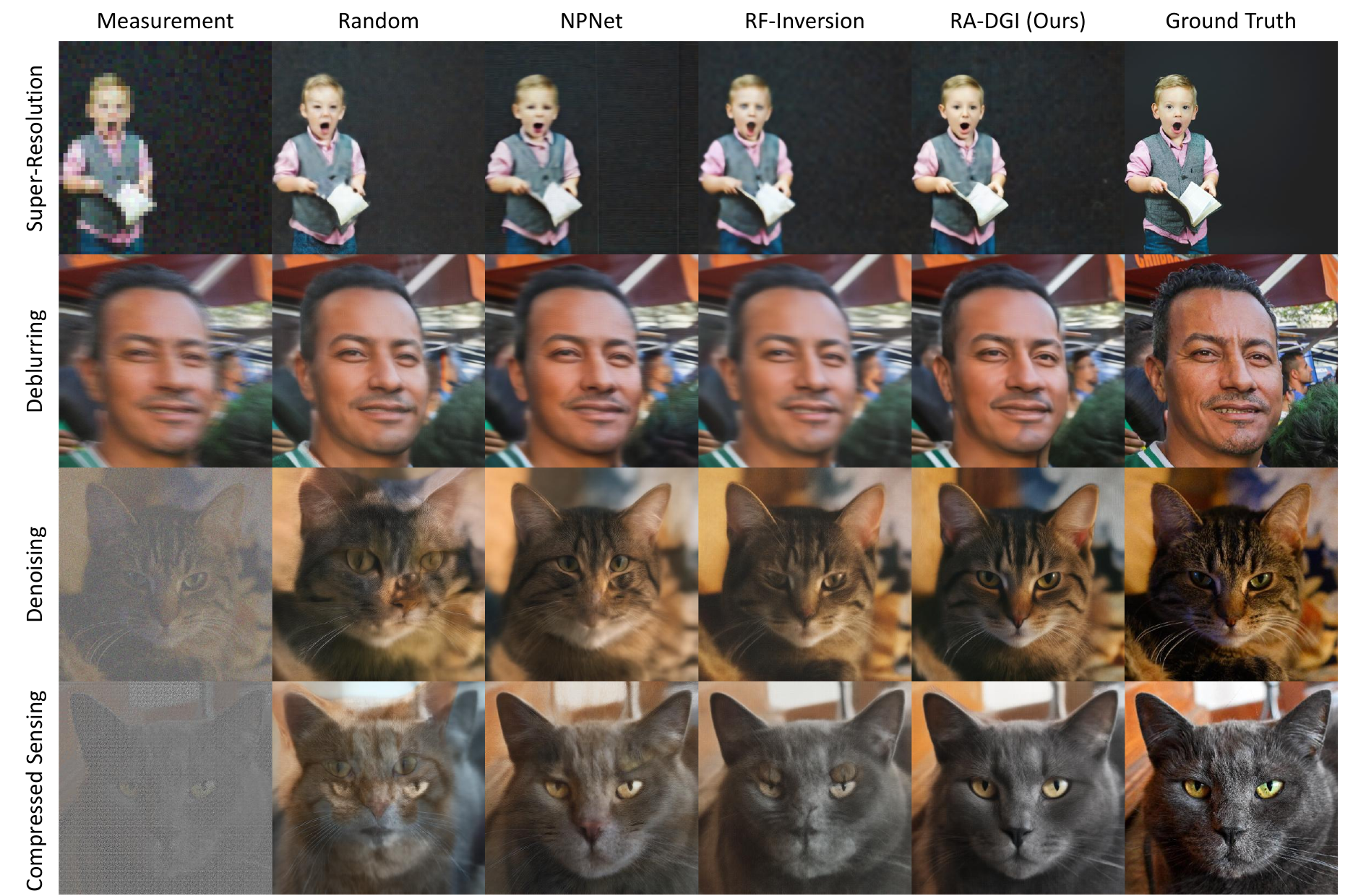}
    \caption{Qualitative comparison on four inverse problems. The displayed examples illustrate improved structural fidelity and detail with RA-DGI}
    \vspace{-3mm}
    \label{fig:qual_comp}
\end{figure*}

\vspace{-5mm}
\paragraph{Comparison Methods}
Our method aims to identify an effective noise initialization strategy for flow-based inverse problem solvers. As a basic baseline, we consider the standard random Gaussian initialization, where the latent noise is sampled directly from the prior distribution. Because initialization strategies for inverse problem solving have been comparatively less studied, we construct several relevant baselines for comparison. First, we include approaches originally proposed for identifying initial noise in image generation, particularly NPNet~\cite{zhou2025golden}. Second, we consider inversion-based strategies that obtain the initial noise by inverting the measurement. In particular, we evaluate RF-Inversion~\cite{rout2025semantic}, an inversion method designed for flow-based models. For a fair comparison, all initialization strategies are integrated into the same inverse problem solver, FlowDPS~\cite{kim2025flowdps}, so that performance differences can be attributed to the initialization quality rather than the solver itself.
\vspace{-4mm}
\paragraph{Implementation Details}
For retrieval-augmented initialization, we use DINOv3~\cite{simeoni2025dinov3} as the feature encoder $\phi(\cdot)$ to extract semantic features for nearest-neighbor search. For the retrieval dataset, we randomly select 1,000 images from Unsplash~\cite{unsplash2020dataset}. All images are resized to $768 \times 768$. During inference, the classifier-free guidance~\cite{ho2021classifierfree} scale is set to 2.0. For AFHQ, we use the text prompt ``a photo of a close-up face of a cat (or dog)''. For FFHQ, we use the prompt ``a photo of a close-up face''. For DIV2K, we generate the text prompt from the degraded image using the degradation-aware prompt extractor (DAPE) introduced in SeeSR~\cite{wu2024seesr}.

\subsection{Main Results}
\vspace{-2mm}
\paragraph{Quantitative Results}
Tab.~\ref{tab:quant_result} summarizes the quantitative comparison across four inverse problems on AFHQ, FFHQ, and DIV2K. For this main comparison, we repeat each experiment using three independent random seeds and report the mean performance, with the standard deviation shown in parentheses. Overall, RA-DGI achieves strong pixel-level reconstruction fidelity across datasets and tasks, attaining the highest mean PSNR in 11 of the 12 settings and the highest mean SSIM in all 12 settings. Improvements in LPIPS and FID vary across settings. Random initialization samples the latent code directly from the Gaussian prior without considering the measurement, which often yields latents inconsistent with the given measurement and thus degrades reconstruction quality. NPNet attempts to identify a more suitable initial noise by aligning the latent with the text prompt. However, because it does not explicitly model the measurement, it often fails to achieve better reconstruction performance than random initialization. RF-Inversion estimates the initial latent by inverting the measurement through the flow model. This strategy partially improves the alignment between the latent code and the target image. However, the inversion process starts from the degraded measurement and does not explicitly enforce measurement consistency during inversion, thereby limiting its performance relative to our approach. In contrast, the proposed RA-DGI, which initializes the inversion from a retrieved proxy clean image and enforces measurement consistency during inversion, results in a latent that is more suitable for solving inverse problems than those produced by the baseline methods.
\vspace{-4mm}
\paragraph{Qualitative Results}
Fig.~\ref{fig:qual_comp} presents qualitative comparisons across four inverse problems. Overall, RA-DGI produces reconstructions that are visually closer to the ground truth than those obtained from other initialization strategies. Random and NPNet initialize the latent without utilizing measurement information and therefore lack structural guidance from the measurement. Consequently, they often produce reconstructions with distorted object structures, particularly in the denoising and compressed sensing results. RF-Inversion incorporates measurement information during initialization and thus restores the overall image structure more reliably than random and NPNet. However, because the latent is obtained by directly inverting the measurement, it can deviate from the Gaussian prior, making it more difficult to generate realistic images. As a result, in tasks such as super-resolution and deblurring, RF-Inversion often fails to accurately recover fine high-frequency details. In contrast, RA-DGI leverages both retrieval-based initialization and dual-guided inversion, enabling it to preserve global structure while recovering realistic high-frequency details.

\subsection{Inference-Time Scaling for Noise Initialization}
Our method selects the initial noise through an inversion procedure, which requires approximately twice the computational time of standard random initialization. Recent work has shown that inference-time scaling can improve diffusion model performance by allocating additional compute at test time~\cite{ma2025scaling}. Motivated by this idea, we investigate whether similar scaling strategies can improve the performance of random noise initialization for inverse problems. We adopt two inference-time scaling strategies, random search and zero-order search, to explore multiple noise candidates during inference. 

\begin{figure*}[ht]
    \centering
    \includegraphics[width=\linewidth]{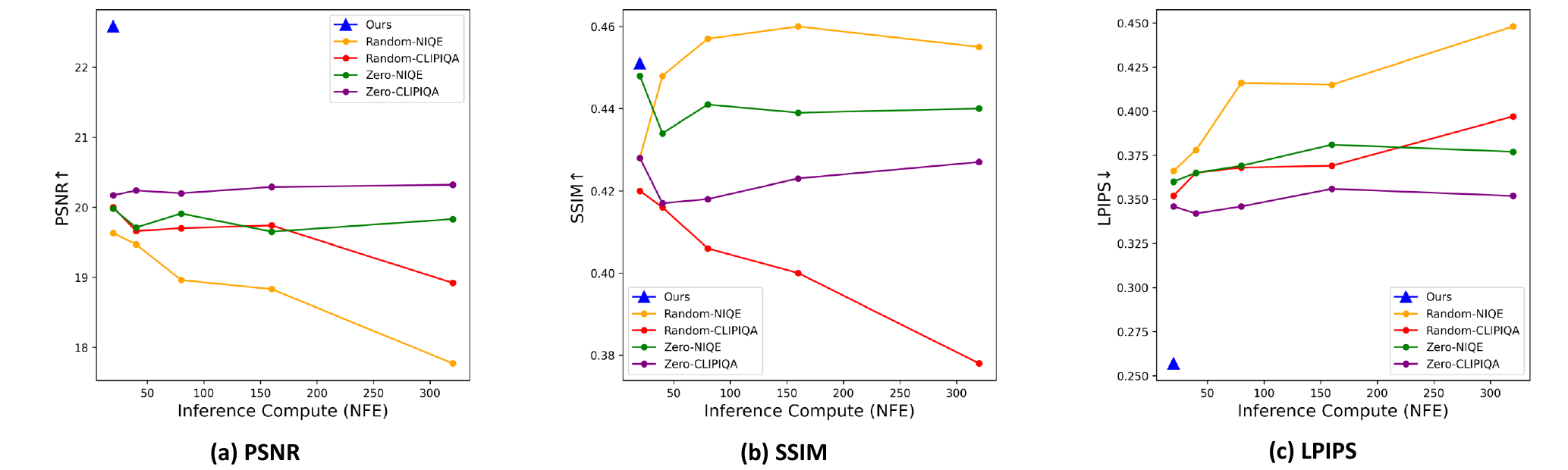}
    \caption{Effect of inference-time scaling compared with our initialization strategy. The x-axis shows inference-time compute (NFE), and the y-axis shows reconstruction metrics (PSNR, SSIM, LPIPS). Although some search configurations achieve higher SSIM, RA-DGI offers a more favorable tradeoff between reconstruction quality and NFE, achieving better PSNR and LPIPS with substantially fewer function evaluations.}
    \label{fig:inf_scale}
\end{figure*}

\begin{table*}[t]
\centering
\scriptsize
\setlength{\tabcolsep}{3pt}
\resizebox{\linewidth}{!}{
\begin{tabular}{c|ccc|cccc|cccc}
\toprule
\multirow{2}{*}{Method} & \multicolumn{3}{c|}{Component} & \multicolumn{4}{c|}{Super-Resolution} & \multicolumn{4}{c}{Deblurring} \\
\cmidrule(lr){2-4} \cmidrule(lr){5-8} \cmidrule(lr){9-12}
 & Retrieval & Prior & Meas. 
 & PSNR$\uparrow$ & SSIM$\uparrow$ & LPIPS$\downarrow$ & FID$\downarrow$
 & PSNR$\uparrow$ & SSIM$\uparrow$ & LPIPS$\downarrow$ & FID$\downarrow$ \\
\midrule

w/o retrieval 
& $\times$ & $\checkmark$ & $\checkmark$
& 23.37 & 0.492 & 0.313 & 31.11
& 22.62 & 0.439 & 0.299 & 31.06 \\

w/o prior guidance
& $\checkmark$ & $\times$ & $\checkmark$
& \textbf{24.52} & \textbf{0.580} & 0.275 & \underline{21.98}
& \textbf{24.13} & \textbf{0.580} & 0.314 & \underline{25.95} \\

w/o measurement guidance
& $\checkmark$ & $\checkmark$ & $\times$
& 23.59 & 0.473 & \textbf{0.231} & 23.63
& 23.22 & 0.470 & \textbf{0.261} & 26.03 \\

RA-DGI (Ours)
& $\checkmark$ & $\checkmark$ & $\checkmark$
& \underline{24.02} & \underline{0.534} & \underline{0.236} & \textbf{21.03}
& \underline{23.57} & \underline{0.525} & \underline{0.267} & \textbf{23.94} \\

\bottomrule
\end{tabular}
}
\caption{Ablation study of the proposed components. We analyze the contributions of retrieval-augmented initialization, prior guidance, and measurement guidance. Best and second-best results are highlighted in \textbf{bold} and \underline{underline}, respectively.}
\vspace{-4mm}
\label{tab:ablation}
\end{table*}

For random search, we scale the number of initial candidates in powers of two, using 2, 4, 8, 16, and 32 candidates. For zero-order search, we use two pivots and two candidates per pivot, and increase the number of optimization iterations from 1 to 5. Since ground-truth images are unavailable at test time, reference-based metrics such as PSNR, SSIM, and LPIPS cannot be used as verification signals. Instead, we employ no-reference image quality assessment metrics, specifically NIQE~\cite{mittal2012making} and CLIPIQA~\cite{wang2023exploring}, as surrogate verifiers.

We conduct experiments on the AFHQ dataset using compressed sensing as the inverse problem. Fig.~\ref{fig:inf_scale} compares our method with random initialization combined with inference-time scaling. The figure reports PSNR, SSIM, and LPIPS results under different scaling strategies, where the legend indicates the scaling algorithm (random search or zero-order search) followed by the verifier used (NIQE or CLIPIQA). When the same amount of inference-time compute is allocated, our method consistently achieves better reconstruction performance across all metrics. Furthermore, for PSNR and LPIPS, increasing the computational budget does not lead to noticeable improvements when inference-time scaling is applied to random initialization. This is likely because the verifiers rely on no-reference image quality metrics rather than ground-truth supervision. Although certain combinations achieve slightly higher SSIM than our method, the additional computational cost required makes our approach more efficient.

\subsection{Ablation Studies}
The proposed RA-DGI consists of three main components: retrieval-augmented initialization, prior guidance, and measurement guidance. To analyze the contribution of each component, we conduct an ablation study by removing one component at a time and comparing the resulting performance. The experiments are performed on the AFHQ dataset and evaluated on two inverse problems: super-resolution and deblurring. The results are reported in Tab.~\ref{tab:ablation}.

In the setting without retrieval, the degraded measurement is used directly as the starting point for inversion, while the same prior- and measurement-guided inversion procedure is retained. The performance degradation observed on both tasks demonstrates that the retrieved proxy provides a more informative starting point than direct inversion of the measurement.

When prior guidance is removed, the influence of measurement guidance becomes stronger. As a result, pixel-level accuracy metrics such as PSNR and SSIM increase. However, the encoded latent deviates further from the Gaussian prior, leading to a degradation in perceptual quality as measured by LPIPS. This observation indicates that the generative prior plays an important role in maintaining realistic image structures during the reconstruction process. Conversely, when measurement guidance is removed, the encoded latent remains closer to the Gaussian prior. This improves perceptual quality, resulting in better LPIPS scores. However, because measurement information is not sufficiently encoded, the pixel-level accuracy decreases. This result underscores the importance of enforcing measurement consistency during inversion for accurate reconstruction. Therefore, both guidance mechanisms are necessary to balance pixel-level accuracy and perceptual quality. Overall, the full RA-DGI model achieves the most balanced performance across both tasks, indicating that the two guidance terms complement each other and jointly improve reconstruction quality.
\section{Conclusion}
In this paper, we investigate initialization strategies for flow-based inverse problem solvers, a topic that remains underexplored in prior work. To address this issue, we proposed RA-DGI, a method for constructing improved initial noise for flow-based inverse solvers. Our approach first retrieves a proxy clean image from a dataset, providing a more suitable starting point for inversion. During inversion, we further introduce dual guidance that jointly enforces measurement consistency and prior regularization, ensuring that the latent trajectory remains both measurement-aware and prior-aligned. Extensive experiments on multiple inverse problems demonstrate that RA-DGI outperforms standard random initialization and existing initialization strategies in overall reconstruction performance. Furthermore, we show that improving initialization is more effective and computationally efficient than inference-time scaling strategies. Overall, our findings highlight initialization as a critical factor in flow-based inverse problem solving. We hope this work encourages further research into initialization strategies for inverse problem solvers.
\clearpage
\section*{Acknowlegement}
This research was supported by Brain Korea 21 FOUR funded by the Ministry of Education of Korea and National Research Foundation of Korea, the Ministry of Science and ICT (MSIT) in Korea under the ITRC support program supervised by the Institute for Information Communication Technology Planning and Evaluation (RS-2020-II201749), and the National Research Foundation of Korea grant funded by the Korea government (RS-2022-00144190).
{
    \small
    \bibliographystyle{ieeenat_fullname}
    \bibliography{main}
}
\clearpage
\setcounter{table}{2}
\setcounter{figure}{4}
\setcounter{equation}{20}

\appendix

\twocolumn[{
\renewcommand\twocolumn[1][]{#1}
\begin{center}
{\LARGE\bfseries
Retrieval-Augmented Initialization with Dual-Guided Inversion for Flow-Based Inverse Problem Solvers\par}
\vspace{0.5em}
{\Large -- Supplementary Material --}
\vspace{1em}
\end{center}
}]

\section{Analysis on Retrieval} \label{sec:appa}
\vspace{-2mm}
We conduct additional analyses of the retrieval module using 100 FFHQ~\cite{karras2019style} images on the super-resolution task. The results are summarized in Tab.~\ref{tab3:retrieval_analysis}.
\vspace{-5mm}
\paragraph{Effect of retrieval dataset size}
We first investigate the impact of the retrieval pool size by varying the number of candidate images used for retrieval. The results are reported in the first block of Tab.~\ref{tab3:retrieval_analysis}. Each experiment is repeated five times, and we report the mean performance together with the standard deviation in parentheses. Overall, the reconstruction quality remains relatively stable as the retrieval pool size increases from 100 to 10,000 images. These results suggest that a large candidate pool is not essential for obtaining effective proxy images. However, when the retrieval pool contains only 100 images, we observe noticeably larger performance variance across repeated runs. This indicates that a small retrieval pool may occasionally fail to provide sufficiently relevant proxy images, leading to a less stable initialization. Considering both performance and computational efficiency, we use a retrieval pool of 1,000 images as the default setting in all experiments.

\vspace{-5mm}
\paragraph{Effect of retrieval encoder}
We next evaluate different feature representations for image retrieval, and report the results in the second block of Tab.~\ref{tab3:retrieval_analysis}. Specifically, we compare DINOv3~\cite{simeoni2025dinov3} features, VAE latent features from Stable Diffusion 3 (SD3)~\cite{esser2024scaling}, and direct retrieval based on measurement similarity. Among all variants, DINOv3 consistently achieves the best reconstruction performance. Replacing DINOv3 with the SD3 VAE encoder leads to noticeable degradation across all evaluation metrics. We attribute this behavior to the different objectives of the two encoders. DINOv3 is trained to capture high-level semantic information and semantic similarity between images, whereas the VAE encoder is primarily optimized for image compression and reconstruction. Consequently, DINOv3 produces more semantically relevant proxy images for initialization. Direct retrieval based on measurement similarity yields the worst performance, indicating that low-level measurement similarity alone is insufficient for identifying semantically meaningful proxy images. These results highlight the importance of semantic-aware retrieval for obtaining effective initialization points. The observed reconstruction metrics are consistent with the motivation for semantic features, but do not directly measure semantic relevance. Direct measurement-similarity retrieval performs worst among the tested representations in this setting; this does not exclude effective low-level retrieval strategies in other settings.
\vspace{-5mm}
\paragraph{Effect of domain gap}
Finally, we analyze the influence of the retrieval dataset domain. The results are reported in the third block of Tab.~\ref{tab3:retrieval_analysis}. We construct retrieval pools using FFHQ (human face dataset), AFHQ~\cite{choi2020stargan} (animal face dataset), and LHQ~\cite{skorokhodov2021aligning} (landscape dataset), which represent progressively larger domain gaps relative to the target FFHQ test set. As expected, retrieval performance degrades as the domain gap increases. Using FFHQ as the retrieval pool yields the best performance since the retrieved proxy images are drawn from the same domain as the target images. When AFHQ is used, the retrieved images still share certain structural characteristics, such as face-centered compositions, resulting in only a modest performance drop. In contrast, LHQ consists primarily of landscape images and exhibits a substantially larger semantic gap from FFHQ. Consequently, the retrieved proxy images become less relevant to the target measurements, resulting in the largest degradation in reconstruction quality. These observations suggest that retrieval quality is closely related to semantic proximity between the retrieval dataset and the target domain.

\begin{table}[t]
\centering
\resizebox{\linewidth}{!}{
\begin{tabular}{c | c | c | c c c}
\toprule
\textbf{Size} & \textbf{Encoder} & \textbf{Dataset} & \textbf{PSNR$\uparrow$} & \textbf{SSIM$\uparrow$} & \textbf{LPIPS$\downarrow$} \\
\midrule

100   & \multirow{3}{*}{DINOv3} & \multirow{3}{*}{Unsplash} & 24.27(0.17) & 0.564(0.018) & 0.206(0.005) \\
1000  & & & 24.24(0.09) & 0.562(0.006) & 0.202(0.006) \\
10000 & & & \textbf{24.39(0.06)} & \textbf{0.572(0.002)} & \textbf{0.198(0.001)} \\
\midrule

\multirow{3}{*}{1000} & DINOv3 & \multirow{3}{*}{Unsplash} & \textbf{24.28} & \textbf{0.571} & \textbf{0.195} \\
 & VAE & & 23.84 & 0.505 & 0.205 \\
 & None & & 23.71 & 0.486 & 0.231 \\
\midrule

\multirow{3}{*}{1000} & \multirow{3}{*}{DINOv3} & FFHQ & 24.05 & \textbf{0.555} & \textbf{0.191} \\
 & & AFHQ & \textbf{24.32} & 0.552 & 0.198 \\
 & & LHQ & 23.84 & 0.513 & 0.200 \\
\midrule

\end{tabular}
}
\vspace{-10pt}
\caption{Retrieval analysis results}
\vspace{-5pt}
\label{tab3:retrieval_analysis}
\end{table}

\begin{table}[t] \centering
\begin{tabular}{lccc}
\toprule & PSNR$\uparrow$ & SSIM$\uparrow$ & LPIPS$\downarrow$ \\
\midrule Shuffled & 23.73 & 0.509 & 0.202 \\
        Nearest neighbor & \textbf{24.28} & \textbf{0.571} & \textbf{0.195} \\
\bottomrule
\end{tabular}
\vspace{-10pt}
\caption{Ablation of the retrieval strategy.}
\vspace{-15pt}
\label{tab:ffhq_sr_retrieval_method}
\end{table}

\section{Additional Ablation Study}
\paragraph{Retrieval method}
To isolate the contribution of semantic retrieval, we compare the nearest-neighbor retrieval used in RA-DGI with random shuffling, where a proxy image is randomly sampled from the dataset. We conduct this comparison on the super-resolution task using 100 FFHQ images. The results are reported in Tab.~\ref{tab:ffhq_sr_retrieval_method}. Nearest neighbor retrieval consistently outperforms random shuffling, indicating that selecting a semantically similar proxy image is important for constructing an effective initialization. This observation is also consistent with the domain gap analysis in Sec.~\ref{sec:appa}, further supporting the use of semantically relevant proxy images.

\paragraph{Inversion method.}
To evaluate the effectiveness of the proposed inversion strategy, we compare the dual-guided inversion used in RA-DGI with ordinary inversion on the super-resolution task using 100 FFHQ images. The results are reported in Tab.~\ref{tab:ffhq_sr_inversion_method}. Dual-guided inversion achieves better performance on most metrics. Consistent with the ablation results in Tab.2, these results indicate that the proposed inversion strategy is important for achieving balanced results in terms of pixel-level fidelity and perceptual quality.

\begin{table}[t]
\centering  \begin{tabular}{lccc}        \toprule        Inversion method & PSNR$\uparrow$ & SSIM$\uparrow$ & LPIPS$\downarrow$ \\        \midrule        Ordinary inversion & \textbf{24.38} & 0.559 & 0.233 \\        Dual-guided inversion & 24.28 & \textbf{0.571} & \textbf{0.195} \\        \bottomrule    \end{tabular}    \caption{Ablation of the inversion method.}    \label{tab:ffhq_sr_inversion_method}
\end{table}

\section{Additional Baseline}
We compare RA-DGI with random initialization under the same nominal budget of 30 NFEs: 30 solve evaluations for Random versus 10 inversion plus 20 solve evaluations for RA-DGI. The experiment is conducted on 100 FFHQ images for the super-resolution task, and the results are reported in Tab.~\ref{tab:ffhq_sr_compute_matched}. RA-DGI achieves slightly better reconstruction performance than random initialization, demonstrating its effectiveness under an equal computational budget.

\begin{table}[t]    \centering    \small    \setlength{\tabcolsep}{4pt}    \begin{tabular}{lccc}        \toprule        Method & PSNR\(\uparrow\) & SSIM\(\uparrow\) & LPIPS\(\downarrow\) \\        \midrule
Random (Solve 30) & 24.55 & 0.618 & 0.182 \\
RA-DGI (Inv. 10 + Solve 20) & \textbf{24.56} & \textbf{0.621} & \textbf{0.178} \\
 \bottomrule    \end{tabular}    \caption{Compute-matched comparison.}    \label{tab:ffhq_sr_compute_matched}\end{table}

\section{Measurement Consistency and Retrieval-Induced Bias}
\label{sec:controlled_initialization}
\paragraph{Measurement consistency}
To evaluate how well the reconstructed images align with the measurements, we measure measurement consistency using $\left|A\hat{x}-y\right|_2^2$, where $\hat{x}$ denotes the reconstructed image. We compute this metric for one seed from the experiments reported in Tab.1, with the results shown in Tab.~\ref{tab:measurement_consistency}. RF-Inversion~\cite{rout2025semantic} and RA-DGI achieve better measurement consistency than the other initialization strategies, as both incorporate the measurement when constructing the initial noise through inversion. RA-DGI further improves measurement consistency over RF-Inversion by explicitly incorporating measurement guidance during the inversion process.

\begin{figure*}[ht]
    \centering
    \includegraphics[width=0.9\linewidth]{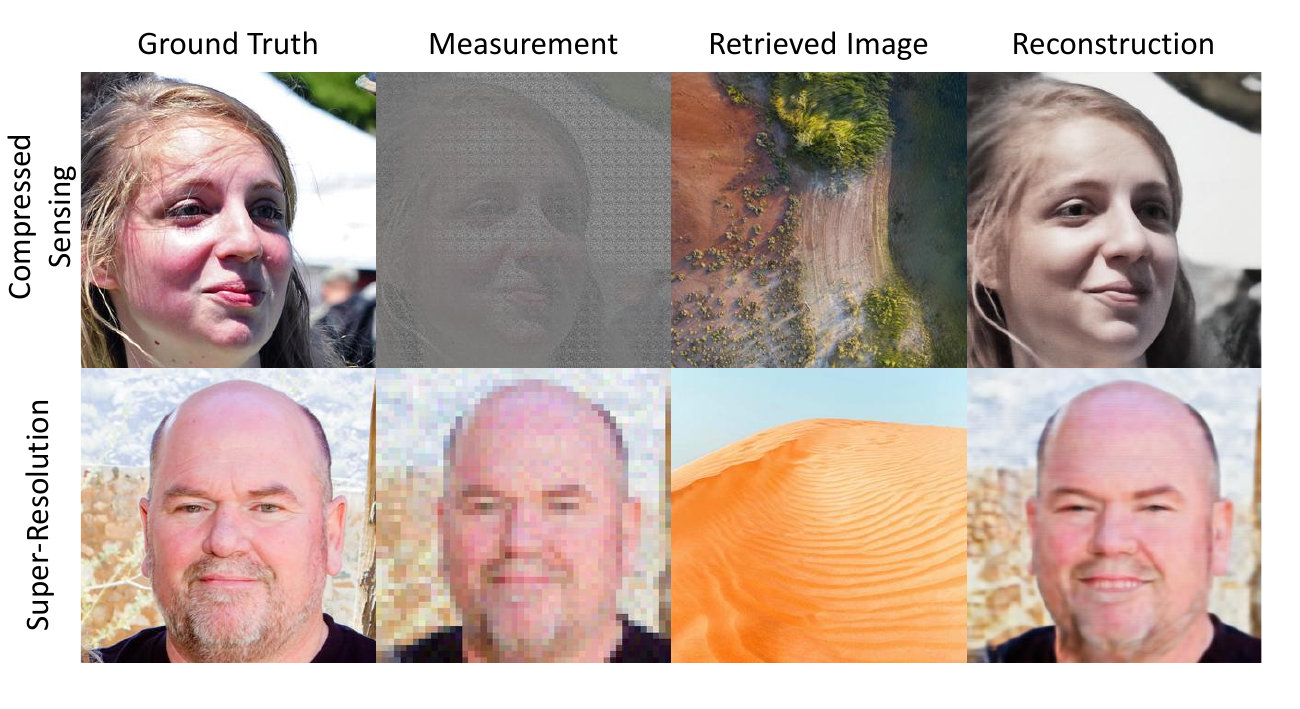}
    \caption{Retrieval-failure cases.}
    \label{fig:retrieval_failure}
\end{figure*}

\paragraph{Retrieval-induced bias}
We qualitatively examine the effect of incorrect retrieval on the final reconstruction. Fig.~\ref{fig:retrieval_failure} shows cases where the retrieved images are semantically unrelated to the ground truth. Despite this mismatch, the reconstructions remain consistent with the measurements and successfully recover the target identity and structure without visibly inheriting the semantics or textures of the retrieved images. We attribute this robustness to the measurement-consistency guidance applied during both inversion and sampling, which continuously anchors the reconstruction trajectory to the observation and limits the influence of irrelevant content from the retrieved image.

\begin{table}[t]    \centering    \scriptsize    \setlength{\tabcolsep}{2.5pt}    \resizebox{\columnwidth}{!}{    \begin{tabular}{llcccc}        \toprule        Dataset & Task & Random & NPNet & RF-Inversion & RA-DGI \\        \midrule        \multirow{4}{*}{AFHQ}        & SR & $0.0055$ & $0.0058$ & $\underline{0.0049}$ & $\mathbf{0.0039}$ \\        & Deblur & $0.0056$ & $0.0059$ & $\underline{0.0055}$ & $\mathbf{0.0036}$ \\        & Denoise & $0.0273$ & $0.0270$ & $\underline{0.0266}$ & $\mathbf{0.0220}$ \\        & CS & $0.0317$ & $0.0411$ & $\underline{0.0177}$ & $\mathbf{0.0154}$ \\        \midrule        \multirow{4}{*}{FFHQ}        & SR & $0.0080$ & $0.0105$ & $\underline{0.0067}$ & $\mathbf{0.0044}$ \\        & Deblur & $0.0081$ & $0.0097$ & $\underline{0.0074}$ & $\mathbf{0.0043}$ \\        & Denoise & $\underline{0.0275}$ & $0.0286$ & $0.0304$ & $\mathbf{0.0214}$ \\        & CS & $0.0313$ & $0.0379$ & $\underline{0.0221}$ & $\mathbf{0.0169}$ \\        \midrule        \multirow{4}{*}{DIV2K}        & SR & $0.0107$ & $0.0111$ & $\underline{0.0080}$ & $\mathbf{0.0064}$ \\        & Deblur & $0.0116$ & $0.0119$ & $\underline{0.0092}$ & $\mathbf{0.0070}$ \\        & Denoise & $0.0527$ & $0.0534$ & $\underline{0.0476}$ & $\mathbf{0.0457}$ \\        & CS & $0.0596$ & $0.0820$ & $\underline{0.0432}$ & $\mathbf{0.0424}$ \\        \bottomrule    \end{tabular}}    \caption{Measurment consistency results.}    \label{tab:measurement_consistency}
\end{table}

\section{Wall-Clock Efficiency}\label{sec:wallclock}
We report the per-image wall-clock latency averaged over 100 images in Tab.~\ref{tab:wallclock}. Except for RA-DGI, the compared methods exhibit similar runtimes. Although RF-Inversion~\cite{rout2025semantic} additionally performs inversion, it relies on ordinary inversion without gradient computation and therefore incurs only a small computational overhead. In contrast, RA-DGI requires an additional inversion process whose runtime is comparable to that of sampling, resulting in approximately twice the latency of random initialization. The additional operations required by RA-DGI, such as retrieval and nearest-neighbor search, are computationally inexpensive and have a negligible impact on the overall runtime. Therefore, we expect the relative trends observed in the NFE-based comparison in Fig.4 to remain largely consistent when evaluated in terms of wall-clock time.

\begin{table}[t]
\centering \small
\begin{tabular}{lc} \toprule Method & Time/image (s)$\downarrow$ \\        \midrule Random & 5.9 \\ NPNet & 6.1 \\ RF-Inversion & 6.3 \\ RA-DGI & 11.2 \\        \bottomrule    \end{tabular}
\caption{Per-image wall-clock latency.}
\label{tab:wallclock}
\end{table}

\section{Compatibility with FLAIR}
\label{sec:flair_compatibility}
To evaluate the compatibility of RA-DGI with existing inverse problem solvers, we integrate its initialization strategy with FLAIR~\cite{erbach2025solving} and evaluate the resulting method on the super-resolution task using 100 FFHQ images. As shown in Tab.~\ref{tab:flair_compatibility}, incorporating RA-DGI consistently improves reconstruction performance over the original FLAIR initialization. These results demonstrate that RA-DGI can be readily combined with different inverse problem solvers to further improve their performance.

\begin{table}[t]
    \centering
    \small
    \setlength{\tabcolsep}{4pt}
    \begin{tabular}{lccc}
        \toprule
        Method & PSNR$\uparrow$ & SSIM$\uparrow$ & LPIPS$\downarrow$ \\
        \midrule
        FLAIR & 23.99 & 0.596 & 0.538 \\
        \textbf{FLAIR + RA-DGI} & \textbf{24.03} & \textbf{0.600} & \textbf{0.532} \\
        \bottomrule
    \end{tabular}
    \caption{Compatibility with FLAIR.}
    \label{tab:flair_compatibility}
\end{table}

\FloatBarrier

\section{Theoretical Analysis}
The following theorem characterizes the probability that a random Gaussian latent lies within a given distance of a fixed target latent.This result provides a quantitative explanation for why a randomly sampled initialization is extremely unlikely to be close to the latent corresponding to a particular image.

\begin{figure*}[ht]
    \centering
    \includegraphics[width=0.9\linewidth]{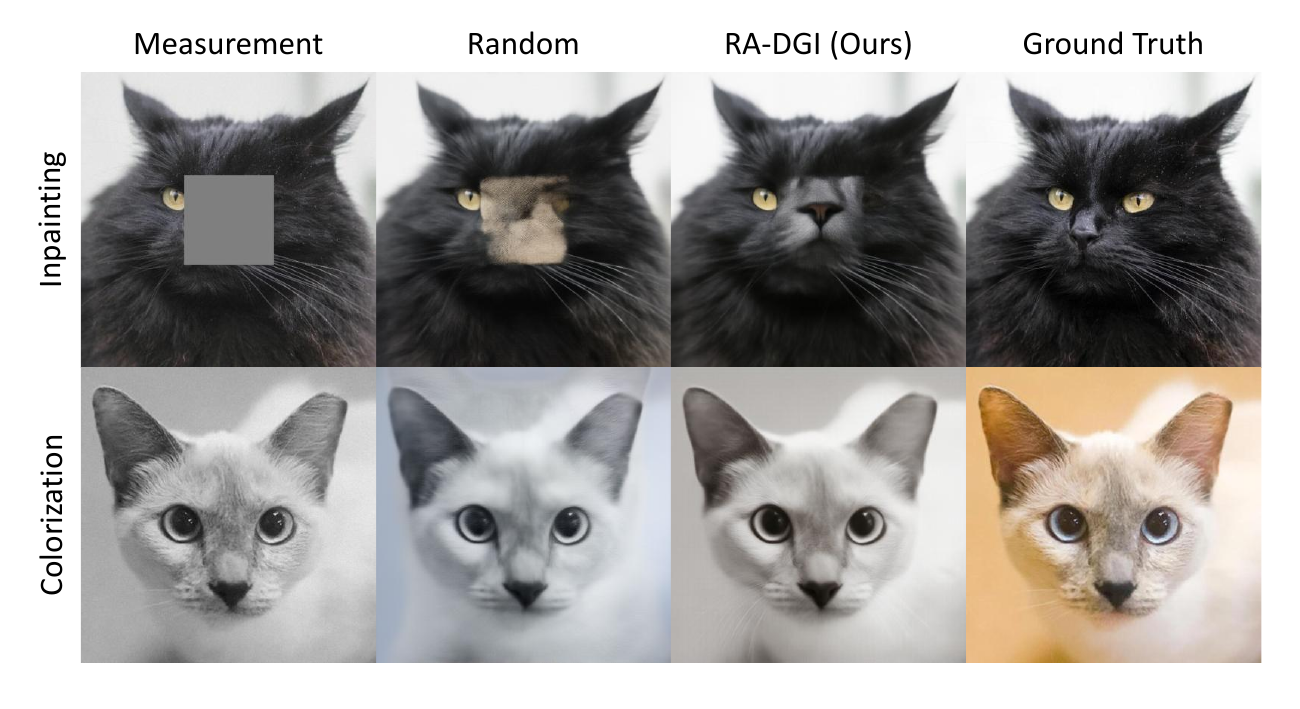}
    \caption{Failure cases on tasks where the underlying solver is weak. Even with the proposed initialization, it remains difficult to overcome the inherent limitations of the original solver.}
    \label{fig:limitation}
\end{figure*}

\begin{theorem}[Distribution of the Distance to a Fixed Latent]
\label{theorem}
Let $z \sim \mathcal{N}(0, I_d)$ be a standard Gaussian latent in $\mathbb{R}^d$, 
and let $z^\star \in \mathbb{R}^d$ be a fixed deterministic latent 
(e.g., the ground truth inversion latent). 
For any radius $r > 0$, define
\[
Y := \|z - z^\star\|_2^2.
\]
Then $Y$ follows a non-central chi-square distribution with
$d$ degrees of freedom and non-centrality parameter 
$\lambda = \|z^\star\|_2^2$, i.e.,
\[
Y \sim \chi'^2_d(\lambda).
\]
Consequently, the probability that a random Gaussian latent lies
within an $L_2$ ball of radius $r$ centered at $z^\star$ is
\[
\mathbb{P}\!\left( \|z - z^\star\|_2 \le r \right)
=
F_{\chi'^2_d(\lambda)}(r^2),
\]
where $F_{\chi'^2_d(\lambda)}$ denotes the cumulative distribution function of the non-central chi-square distribution.
\end{theorem}

\begin{proof}
Let $z = (z_1, \dots, z_d)^\top \sim \mathcal{N}(0, I_d)$ and 
let $z^\star = (z^\star_1, \dots, z^\star_d)^\top \in \mathbb{R}^d$ 
be a fixed deterministic vector. Define
\[
Y = \|z - z^\star\|_2^2
= \sum_{i=1}^d (z_i - z^\star_i)^2 .
\]
Since $z_i \sim \mathcal{N}(0,1)$ independently, each shifted variable
\[
z_i - z^\star_i \sim \mathcal{N}(-z^\star_i,\,1).
\]
Therefore, $(z_i - z^\star_i)^2$ is the square of a Gaussian random variable with mean $-z^\star_i$ and unit variance. By definition, the sum of squares of independent Gaussian variables with means $\mu_i$ and unit variances follows a non-central chi-square distribution with $d$ degrees of freedom and non-centrality parameter
\vspace{-3mm}
\[
\lambda = \sum_{i=1}^d \mu_i^2 .
\]
In our case $\mu_i = -z^\star_i$, hence
\[
\lambda = \sum_{i=1}^d (z^\star_i)^2 = \|z^\star\|_2^2 .
\]
Consequently,
\[
Y = \sum_{i=1}^d (z_i - z^\star_i)^2
\sim \chi'^2_d(\lambda),
\]
where $\chi'^2_d(\lambda)$ denotes the non-central chi-square distribution
with $d$ degrees of freedom and non-centrality parameter $\lambda$.
Finally,
\[
\mathbb{P}(\|z - z^\star\|_2 \le r)
=
\mathbb{P}(Y \le r^2)
=
F_{\chi'^2_d(\lambda)}(r^2),
\]
which completes the proof.
\end{proof}

Theorem~\ref{theorem} shows that the distance between a random Gaussian latent and a fixed target latent follows a non-central chi-square distribution. Therefore, the probability that a random initialization falls within any prescribed neighborhood of the target latent can be computed directly from the corresponding cumulative distribution function.

\paragraph{Remark 1 (Typical norm of Gaussian latents).}
For $z \sim \mathcal{N}(0,I_d)$, define $S := \|z\|_2^2$. Then $S$ follows a chi-square distribution with $d$ degrees of freedom, i.e., $S \sim \chi_d^2$, with mean $\mathbb{E}[S]=d$ and variance $\operatorname{Var}(S)=2d$. Although the standard deviation of $S$ grows as $\sqrt{2d}$, its mean grows as $d$. Therefore, its relative standard deviation, $\sqrt{\operatorname{Var}(S)}/\mathbb{E}[S]=\sqrt{2/d}$, decreases as the dimension increases. Consequently, deviations of $\|z\|_2^2$ from its mean become small relative to $d$, supporting the approximation $\|z\|_2^2 \approx d$ for a typical Gaussian latent in high dimensions.

\subsection{Numerical Example}
We use Theorem~\ref{theorem} to illustrate the proximity advantage of RA-DGI initialization over random Gaussian initialization. Let $z^\star$ denote the latent obtained by inverting the ground truth image, and let $z \sim \mathcal{N}(0,I_d)$ denote a randomly sampled Gaussian latent. Theorem~\ref{theorem} characterizes the distribution of their squared Euclidean distance, allowing us to compute the probability that $z$ lies within a given radius of $z^\star$.

In our experiments, the mean Euclidean distance between the ground truth latent and the RA-DGI initialization is approximately $516$ across 100 evaluation samples. We therefore set $r=516$ as a representative radius in Theorem~\ref{theorem}. To specify the non-centrality parameter, we assume that sufficiently accurate inversion maps ground truth images approximately to the model's standard Gaussian prior, and thus regard $z^\star$ as a typical Gaussian latent. The SD3~\cite{esser2024scaling} latent space has dimensionality $d=96\times96\times16=147{,}456$. At this dimensionality, the norm concentration described in Remark~1 motivates the approximation $\lambda=\|z^\star\|_2^2 \approx d$.

Using Theorem~\ref{theorem} with these parameters, the probability that a randomly sampled Gaussian latent lies within this representative distance of the target is
\[
\mathbb{P}\!\left(\|z-z^\star\|_2 \le 516\right)
\approx
F_{\chi'^2_{d}(d)}(516^2)
\approx 1.27 \times 10^{-216}.
\]
This result shows that a random Gaussian initialization is highly unlikely to match RA-DGI’s average proximity to the target latent, supporting RA-DGI’s proximity advantage.

\section{Limitation}
Our method focuses on improving the initial noise rather than modifying the inverse problem solver itself. In this work, experiments are conducted using the FlowDPS~\cite{kim2025flowdps} solver. Therefore, the overall reconstruction performance is inherently bounded by the capability of this solver. Previous studies have shown that FlowDPS performs well on tasks that preserve sufficient structural information, such as super-resolution and deblurring. However, it struggles with problems where the measurement provides extremely limited information, such as inpainting and colorization. Fig.~\ref{fig:limitation} illustrates such cases. In the inpainting example, even with the proposed initialization, the missing region still contains inaccurate content. Similarly, in the colorization example, both random initialization and the proposed method struggle to produce colorized results that faithfully match the ground truth. These results indicate that improving initialization alone cannot fully overcome the solver's intrinsic limitations when the measurement is highly ambiguous.

\section{Additional Results}
\begin{figure*}[ht]
    \centering
    \includegraphics[width=0.9\linewidth]{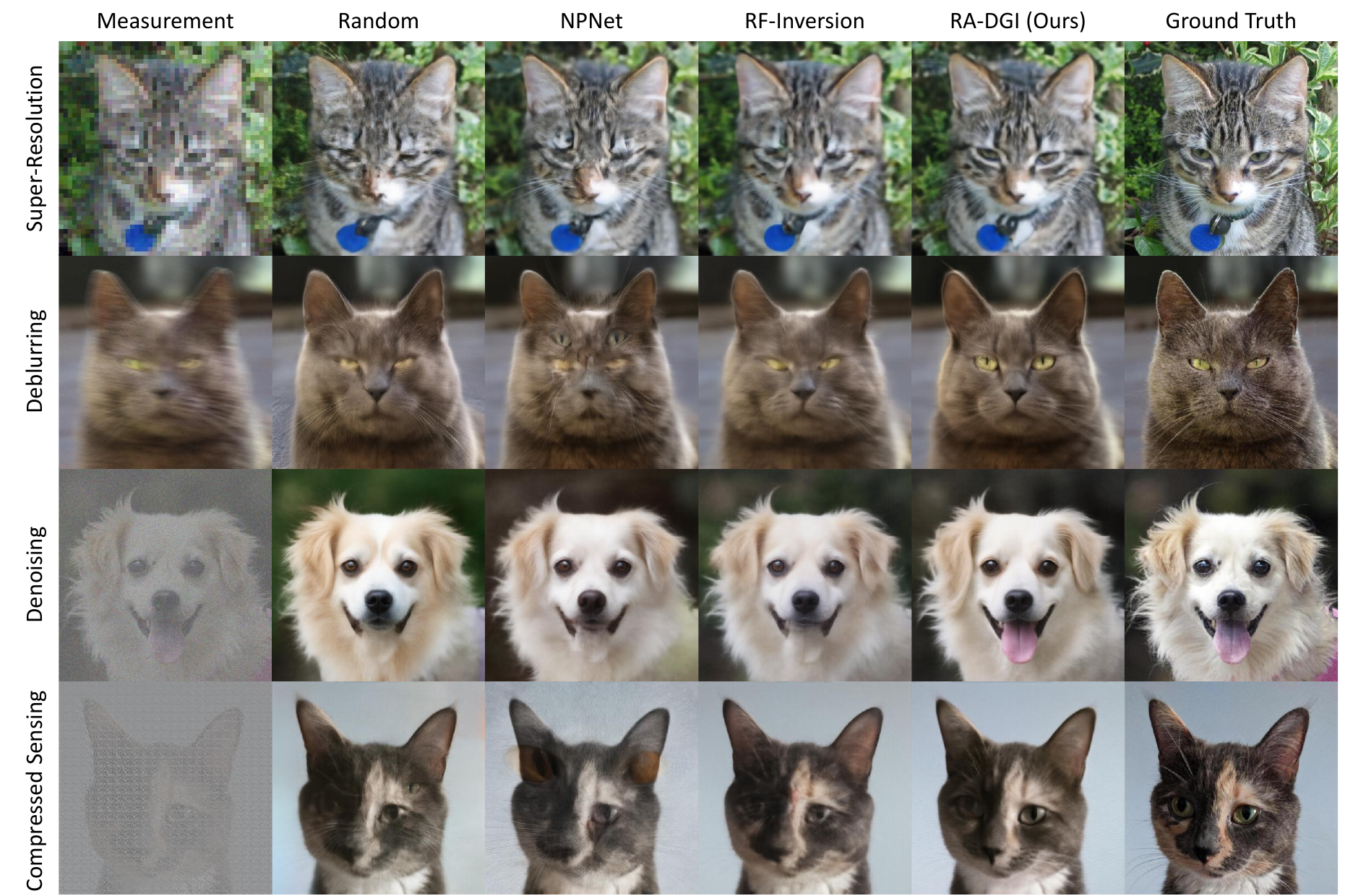}
    \caption{Additional qualitative comparisons on the AFHQ dataset.}
    \label{fig:add_afhq}
\end{figure*}

\begin{figure*}[ht]
    \centering
    \includegraphics[width=0.9\linewidth]{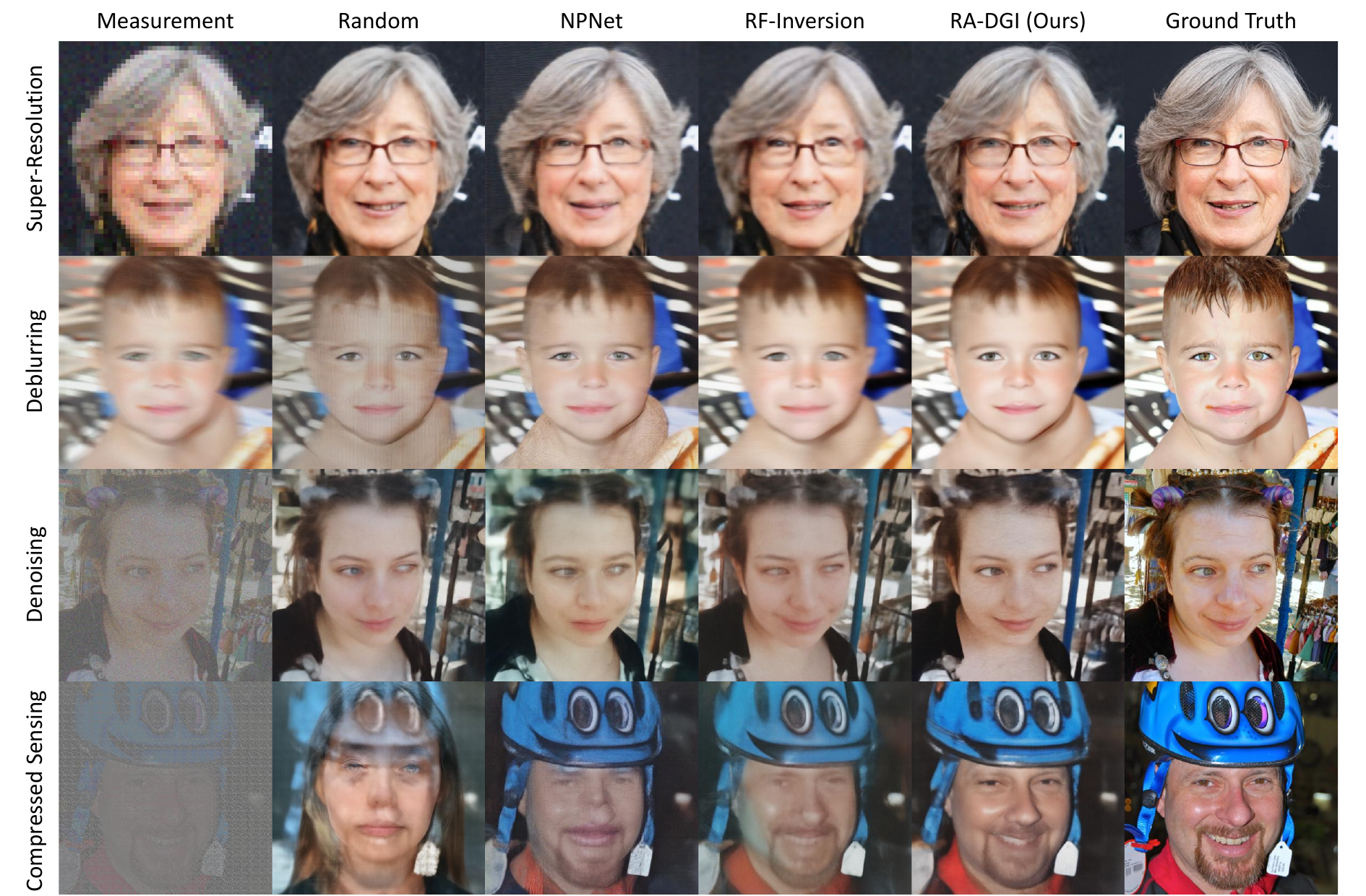}
    \caption{Additional qualitative comparisons on the FFHQ dataset.}
    \label{fig:add_ffhq}
\end{figure*}

\begin{figure*}[ht]
    \centering
    \includegraphics[width=0.9\linewidth]{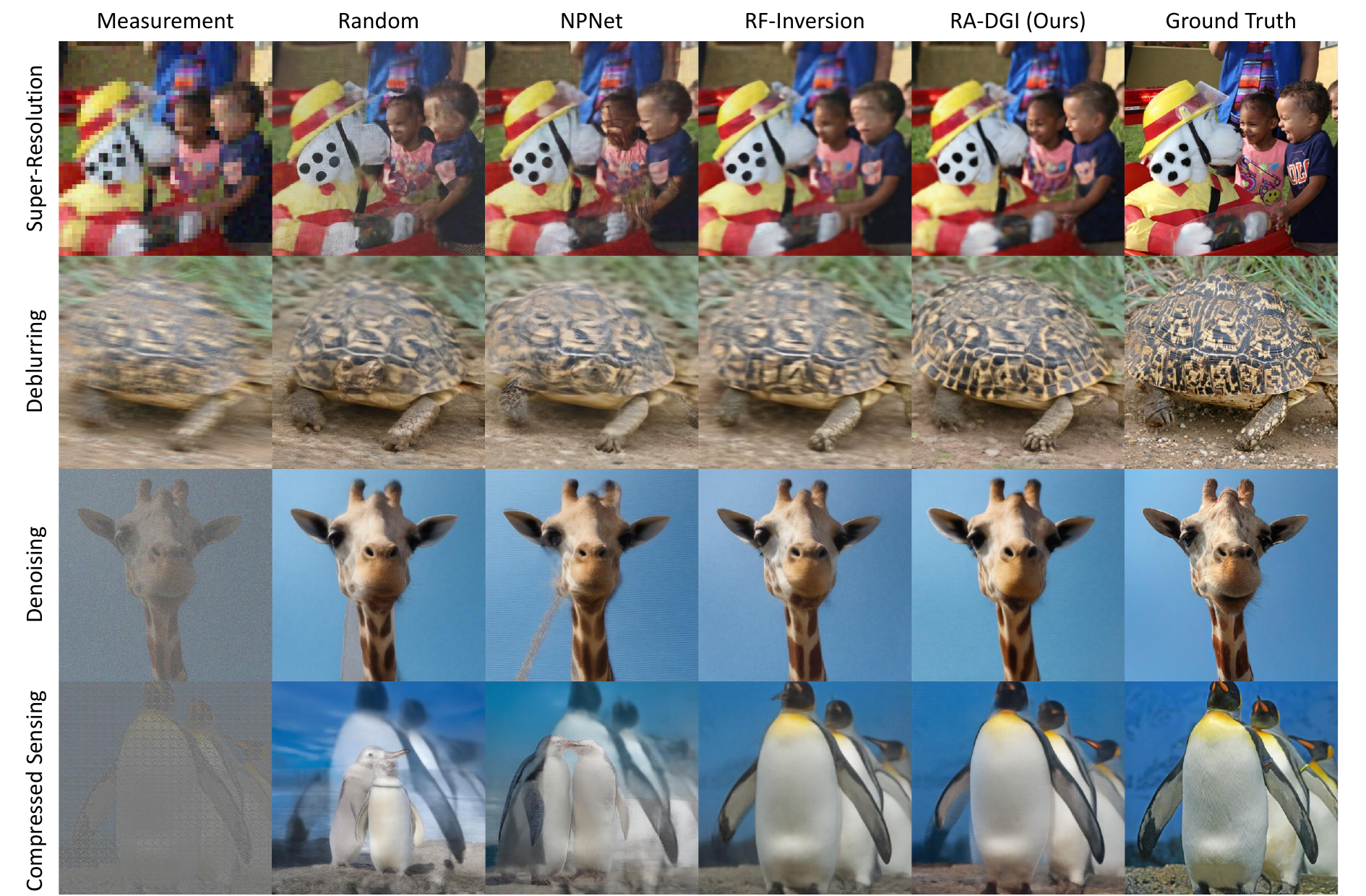}
    \caption{Additional qualitative comparisons on the DIV2K dataset.}
    \label{fig:add_div2k}
\end{figure*}

We provide additional qualitative comparisons across three datasets. Fig.~\ref{fig:add_afhq}, Fig.~\ref{fig:add_ffhq}, and Fig.~\ref{fig:add_div2k} present representative results on the AFHQ, FFHQ, and DIV2K datasets, respectively. Across different datasets and degradation settings, the proposed initialization consistently produces reconstructions that are visually closer to the ground truth, while reducing structural distortions and unrealistic artifacts.


\end{document}